\documentclass[conference]{IEEEtran}
\IEEEoverridecommandlockouts
\usepackage{cite}
\usepackage{amsmath,amssymb,amsfonts}
\usepackage{algpseudocode}
\usepackage{algorithm}
\usepackage{graphicx}
\usepackage{subcaption}
\usepackage{textcomp}
\usepackage[hidelinks]{hyperref}
\usepackage{xcolor}
\def\BibTeX{{\rm B\kern-.05em{\sc i\kern-.025em b}\kern-.08em
    T\kern-.1667em\lower.7ex\hbox{E}\kern-.125emX}}
\begin{document}

\title{On Unsupervised Image-to-image translation and GAN stability}

\author{\IEEEauthorblockN{BahaaEddin AlAila, Zahra Jandaghi, Abolfazl Farahani and Mohammad Ziad Al-Saad}
\IEEEauthorblockA{\textit{Department of Computer Science} \\
\textit{The University of Georgia}\\
Athens, GA, 30606 \\
\href{mailto:bahaaeddin.alaila@uga.edu}{bahaaeddin} 
\href{mailto:zahra.jandaghi@uga.edu}{zahra.jandaghi}
\href{mailto:a.farahani@uga.edu}{a.farahani}
\href{mailto:mohammad.alsaad@uga.edu}{mohammad.alsaad}  @uga.edu
}
}

\maketitle

\begin{abstract}
The problem of image-to-image translation is one that is intruiging and challenging at the same time, for the impact potential it can have on a wide variety of other computer vision applications like colorization, inpainting, segmentation and others. Given the high-level of sophistication needed to extract patterns from one domain and successfully applying them to another, especially, in a completely unsupervised (unpaired) manner, this problem has gained much attention as of the last few years. It is one of the first problems where successful applications to deep generative models, and especially Generative Adversarial Networks achieved astounding results that are actually of real-world impact, rather than just a show of theoretical prowess; the such that has been dominating the GAN world. In this work, we study some of the failure cases of a seminal work in the field, CycleGAN\cite{60} and hypothesize that they are GAN-stability related, and propose two general models to try to alleviate these problems. We also reach the same conclusion of the problem being ill-posed that has been also circulating in the literature lately.
\end{abstract}

\section{Introduction}

Many image processing and computer vision tasks can be considered as image-to-image translation in which the representation of an object or scene is converted or coerced into another. In the past few years, these tasks have been done by supervised learning methods which require a large number of labeled data examples of matching image pairs. In practice, the availability of such datasets is very scarce and most of the time just plain non-existent. E.g., for the problem of translating an image of a horse to one of a zebra, a scenery of a horse and the exact same scenery but with a zebra with the exact same pose is required. And a whole dataset of those. Thus, the task of unsupervised image-to-image translation is getting the attention of computer vision research community, since it is a thought-provoking problem that has a very significant impact in computer vision applications.

Unsupervised image-to-image translation is a method in which a joint distribution of images in various domains is learned from the marginal distributions of individual domains. These types of unsupervised methods usually result in a large set of joint distributions in which some of them may have no relation to given marginal distribution without additional assumptions or criteria. Therefore, this problem could be addressed as learning the conditional distribution of corresponding images in the target domain, given an image in the source domain \cite{11}. This task could help in altering several aspect of a given image to another such as changing the expression of a person from frowning to smiling \cite{12}. Additionally, could help in solving many problems in the context of computer vision such as inpainting, colorization, segmentation, and increasing the resolution of a given image without quality degradation (super-resolution)\cite{11}. In colorization, the problem could be addressed as a mapping of scaled gray images to a corresponding color image. Also, the super resolution problem could be considered as translating a low-resolution image to a high resolution image. In this work, we focus on using the unsupervised setting where the problem could be harder and more challenging. this is because of two major reasons. First, the process of collecting aligned training example pairs usually do not exist or are very hard to collect, therefore the only evaluation of the model is empirical and therefore, not covered by the generalization theorem, therefore not guaranteed to work out of sample. Second, many mappings are multimodal (many-to-many), where an input image could correspond to diverse possible outputs \cite{35}.

In the recent years, generative adversarial networks (GAN) \cite{21} have increasingly become the research interest of machine learning and artificial intelligence researchers. The idea of GAN was inspired by the zero-sum game, in which each player’s gain or loss is the opposite as the gain or loss of the other player, and the sum of the two players utility is zero. GANs consist of a generator and a discriminator that are trained under the adversarial learning idea. The purpose of the generator is to try to absorb the probable distribution of the real samples, then, to generate new data samples. The discriminator is often a binary classifier with where its goal is to recognize the distinctions between real samples from generated samples as accurately as possible. The main goal of GANs is to estimate the underlying distribution of the real data samples and produce new samples from that distribution.

Recent works that tackle the image-to-image translation problem consider that there is some relationship between the two domains. For instance, CycleGAN \cite{60} was built on the assumption of the presence of an inverse mapping $F$ that translates from $Y$ to $X$ and on cycle-consistency. They train two generators which are bijections and inverse to each other and uses adversarial constraint to ensure the translated images seem to be drawn from the target domain. The cycle-consistency constraint is to ensure the translated image can be mapped back to the original image using the inverse mapping ($F(G(x)) \approx x$ and $G(F(y)) \approx y$). However, UNIT \cite{36} assumes a shared-latent space, meaning a pair of images in different domains can be mapped to some shared latent representations. The model trains two generators $G_X$ and $G_Y$ with shared layers. Both $G_X$ and $G_Y$ map an input to itself, while the domain translation is realized by letting $x_i$ go through part of $G_X$ and part of $G_Y$ to get $y_i$. The model is trained with an adversarial constraint on $G_X$ and $G_Y$, another cycle-consistency constraint, and a variational constraint on the latent code. Assuming cycle-consistency is assumed to ensure 1-1 mapping and avoids mode collapses, and both models generate reasonable image translation and domain adaptation results. Nevertheless, there are some problems with such methods. First, cycle-consistency does not assure that the mapping learned is the anticipated mapping. Theoretically, CycleGAN could find any random 1-1 mapping that fulfills the constraints. Having several global optima is problematic since, in our experiments, we observed that the training is far from stable, meaning it does not guarantee to converge or reproduce the same results every time we redo the training. Also, there is a sensitive trade-off between how similar the translation resembles the target domain, the correctness of the translated image to the input image, and the need for extreme manual tweaking of the weight between the reconstruction loss and the adversarial loss to get sustaining outcomes. Moreover, most of the time we only care about one-way translation, though CycleGAN always requires the training of two generators that are bijections. This not only is cumbersome but it is also hard to balance the effects of the two generators and two discriminators.

Because of their training instability and high tendency for mode collapses, we hypothesize that there is no need for two-GAN parts for an unsupervised image-to-image translation model. Instead we think that having just one good and reliable GAN formulation can achieve the same results, while being less prone to instabilities and mode collapses. The proposed 1-GAN model does not take into consideration the assumption of shared representation or double cycle consistency (for each domain), in which it learns two-way mapping, by training for cycles in just one direction (A->B->A, without B->A->B). Moreover we propose another formulation to solve the translation problem that is GAN-free, by leveraging the probabilistic autoencoding capabilities of variational autoencoders \cite{33} without relying on GANs.

However, as our experiments progressed it became clear that the unsupervised image-to-image translation problem, if under-constrained, is ill-posed and many arbitrary mappings that abide by the cycle constraints are possible. This problem is generally known as the Manifold Alignment problem. The problem is more obvious in our GAN-free formulation since the representation of the images are simple vectors where any and all convolution locality is lost.

\section{Related Work}

\noindent\textbf{Simulation} Several computer vision problems are known as image to image translation, where mapping an image from one domain corresponds to another image from another domain. The progress in graphic synthesis led the researchers to learn the models on synthetic images to reduce the need of annotation which are expensive to produce. However, there is a gap between synthetic and real image distributions which prevents the learning process from performing well. To tackle this problem SimGAN\cite{49} proposed a simulated and unsupervised learning, where it aims to enhance the realism of synthetic images from a simulator using unlabeled real data. CoGAN\cite{37}, coupled generative adversarial networks, is an unpaired image-to-image translation model that learns a joint distribution of multi-domain images without any tuple of corresponding images. This is done by using a weight-sharing strategy as a constraint to the layers that decode the abstract semantics which learns a common representation across domains.

\noindent\textbf{Cycle Consistency} Unlike the aforementioned methods, CycleGAN\cite{60} does not depend on any task-specific, predefined similarity functions between the input and output, or the assumption that input and output have to lie in the same low-dimensional embedding space. This model uses cycle consistency loss and can learn to capture special features in one image domain and translate them into the other image domain while there is no paired training sample. Although the approach performs well when the tasks involve only color and texture changes, little success is achieved when dealing with geometric changes. The DualGAN\cite{57} and DiscoGAN\cite{32} share the same idea as CycleGAN, however all of these models suffer from limitations caused by the assumption the underlying interdomain is deterministic and that only one to one mappings could be learnt. The problem of the above models are dealing with the ill-posed problem meaning that there is no direct constraint on the translated images to map the input image to a corresponding similar looking output image other than the cycle consistency which is a very weak constraint and can be satisfied with arbitrary mapping.

\begin{figure*}
    \centering
    \includegraphics[width=0.55\textwidth,height=5cm]{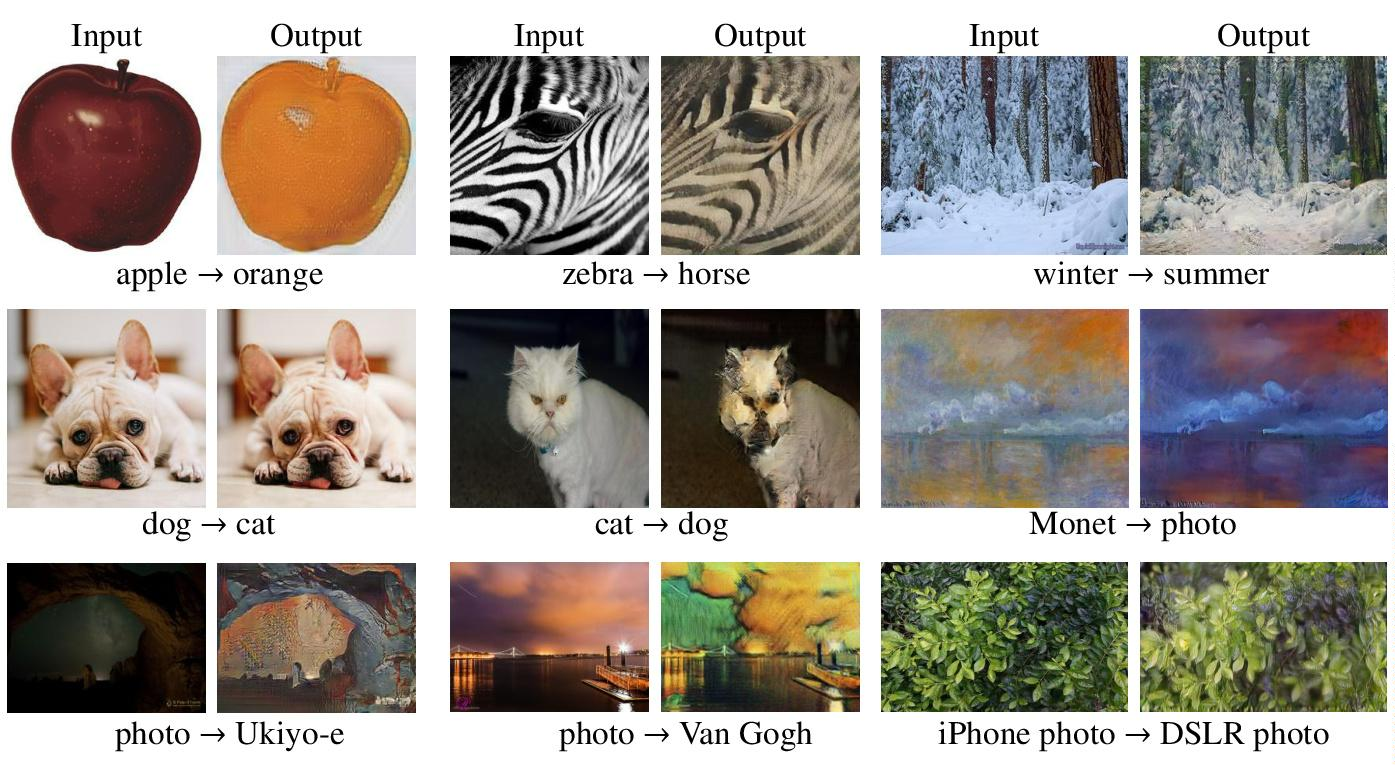}
    \includegraphics[width=0.40\textwidth,height=5cm]{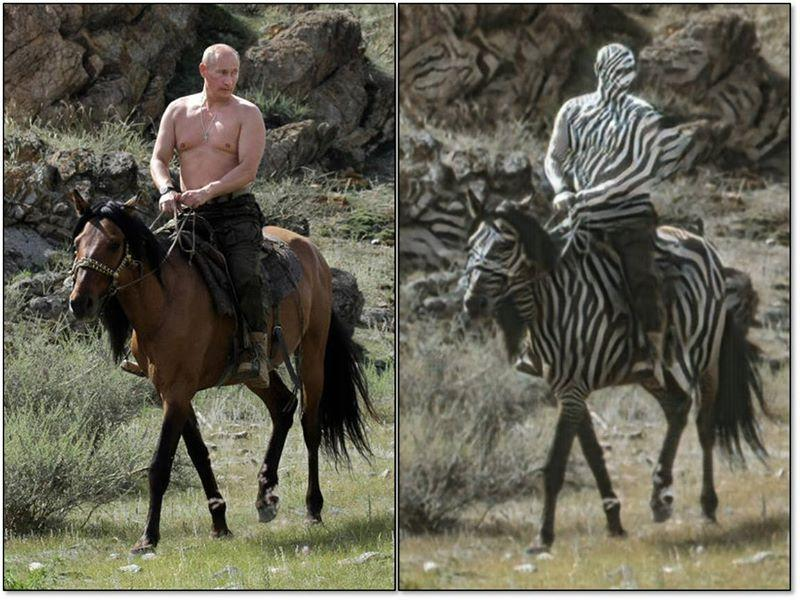}
    \caption{Failure instances reported by CycleGAN\cite{60}}
    \label{fig:1}
\end{figure*}

\noindent\textbf{Multimodality} To tackle this issue, Augmented CycleGAN\cite{5} was proposed claiming that the model could specifically augment each domain with supplementary latent variables and expand CycleGAN training procedure to the augmented spaces, thus providing more constraints to the mapping problem. The mappings in the augmented space could be described as one to one mapping, but many to many in the original domains after marginalizing over the latent variables. UNIT\cite{36}, is the extended version of Coupled GAN and was proposed based on the assumption of making a shared latent space that assumes a pair of corresponding images in different domains that could be mapped to a same latent representation in a shared latent space. This model is built on variational autoencoders\cite{33} and generative adversarial networks where each image domain is VAE-GAN and the adversarial training objective interacts with a weight-sharing constraint, which enforces the shared latent space to produce corresponding images in the respective domains. MUNIT\cite{28} is also proposed to tackle the limitation of one-to-one mapping and generate multiple outputs from a given source domain image. First, this model assumes that image representation or latent space of image can be decomposed into a content space, content code, which is domain-invariant and can be shared between the images between different domains, and a style space, style code, which is specific properties in a domain. Then, to translate an image from one domain to another domain, they recombine its content code, the encoded information that should be preserved during the translation, with a random style code, the encoded information for the remaining variations in the target style space that are not contained in the input image. This decomposition lead the model to generate high quality and diverse translated images. BicycleGAN\cite{61} is the combination of cVAE-GAN and cLR-GAN to jointly makes a connection between latent encoding in both direction in order to generate realistic and more diverse results across wide range of image-to-image translation problems.

\noindent\textbf{Manifold Alignments} if the problem is left completely under-constrained, it results in the problem of aligning two completely alien manifolds. This is realized both by Augmented CycleGANs\cite{5} and MAGAN\cite{6}. Both works argue that in the case of under-constraining, a consistent mapping can only be guaranteed by providing a handful of matched pairs in the domains, such that the generators learn to map the corresponding features for those supervised pairs, and as a result biasing the whole translation process.

Task semantics CyCADA\cite{26} offers the interpretability by visualizing the intermediate output of the method and directs transfer between domains according to a defined discriminatively trained task and escapes divergence by imposing consistency of the relevant semantics before and after adaptation.

Attention mechanisms There are several works incorporating attention mechanism on GANs to make the changes in the translated images as minimal as possible. In this family of models, Self-Attention GAN\cite{59} employs non-local neural network incorporated into GAN in order to model long-range dependencies and enabling both the generator and the discriminator to efficiently model relationships between widely separated spatial regions, so that details could be simply generated using cues from all feature locations. DA-GAN\cite{38} is another example of attention GAN which decomposes the task into translating instances in a highly-structured latent space by feeding image into a localization function and finding some attention areas, then translating the attentions to latent space representations for both domains. This model applies constraints on both instance-level and set-level to generate more accurate translation. Similar to the above approaches, Attention-GAN\cite{11} aims to transform a specific regions or objects in an image to another objects without altering the other irrelevant parts of the image. It decomposes GAN into two parts, the attention network and the transformation network. The attention network focuses on the regions of interest in an image to predict the spatial attention maps while suppressing background and the transformation network translate the objects from source domain to the target domain. Also, \cite{3} tends to focus on specific objects or regions in an image without changing other parts of image by incorporating UNIT in which the discriminator learns accurate maps with no additional supervision. They add an attention network to each generator in the CycleGAN setup in which they are jointly trained to produce attention maps for the regions that are discriminative between the source and target domains and leverage the discriminator to learn this mapping.

Models elaborating on cycle consistency try to provide more constraints to the translation problem one way or another such that the translation training process is more stable and predictable.

\section{Methodology}

\subsection{Motivation}

While studying the pathological image translation cases reported by CycleGAN\cite{60}, we hypothesized that those failure instance (Fig \ref{fig:1}) were, among other things, GAN-related.

GANs are known to be prone to mode collapses\cite{45, 7}, where the Generator of the GAN focuses on only some aspects of the target dataset while ignoring others. This is because the discriminator does not ask whether the generator is hitting all the modes (variations) in the target domain, but only tests whether a generated instance passes as an instance from the target domain or not. Therefore, in the extreme case, if the generator only generates just one real-looking target image unconditional from the generator’s input, the discriminator will have no qualm with it, and hence the generator will not be penalized for lack of diversity. This is true for the original GAN and its variants, which what CycleGAN is using.

Furthermore, the discriminator is also just a binary classifier; it relies on uncovering distinguishing features between the generated instances and the target datapoints. The generator’s goal is to fool the discriminator, i.e., through back-propagating through the discriminator, the generator learns to generate instances that have such target-domain distinguishing features, and move away from previous generator-domain distinguishing features. Therefore if the discriminator does not do a good job on extracting very reliable, domain-defining features for the target domain, the generator will simply not learn to generate very target-looking instances, but only instances that satisfy the superficial features that the discriminator uncovered. Thus, it is imperative to have a sophisticated discriminator that is able to extract sophisticated features, and allow for a lengthy-enough training, since sophisticated models usually take more time to optimize.

However, since the target dataset is finite, overtraining the discriminator can have the adverse effect of overfitting to the target dataset instances, rejecting any other instances even if they empirically look like the target domain. Hence combating overfitting is essential for a successful GAN training, and is something that is touched on in \cite{18}.

It was imperative to us that we should either improve the GAN condition or find another way to model the source and target domain distributions without the use of GANs. Therefore, we propose two general formulations:

\begin{itemize}
\item \textbf{A 1-GAN model}: This is a smaller version of CycleGAN; one encoder, one decoder, and just one discriminator.
\item \textbf{A GAN-free model}: here we rely on variational autoencoders to model the source and target domain distributions, and enforce a cycle consistency to carry out the cross-domain translation
\end{itemize}

\subsection{ The 1-GAN Model}

\begin{figure}
    \centering
    \includegraphics[width=0.48\textwidth]{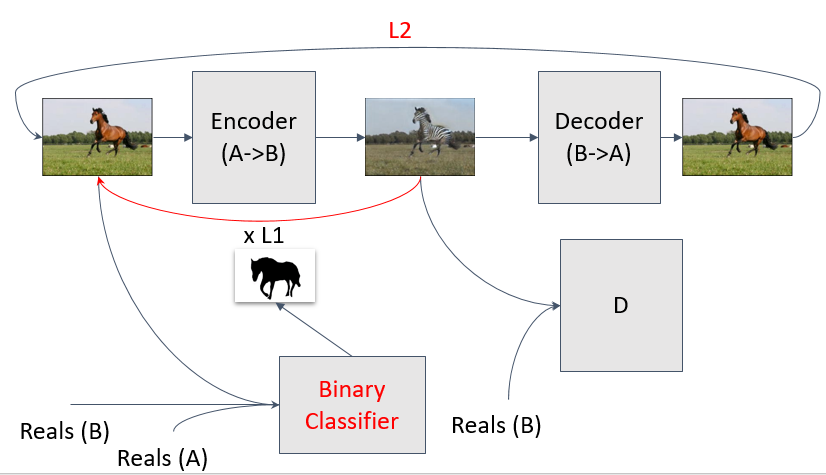}
    \caption{The high-level structure of our 1-GAN model}
    \label{fig:2}
\end{figure}

The encoder and decoder in our 1-GAN model correspond to CycleGAN’s two generators, i.e., the both accept images and generate out images.The discriminator acts on the encoder, and pushes the output of the encoder to be of a the opposite domain of the encoder’s input. The decoder translates the other way around. In other words, given images from domains $A$ and $B$, the encoder shall translate an image of $A$ to an image of $B^{'}$ while the decoder’s job is to translate the $B^{'}$ image back to $A$. The discriminator tests whether the image $B^{'}$ looks similar to images from $B$. In order to constrain the problem more, we propose have a similarity constraint between the input image ($A$) and the translation output ($B^{'}$). The similarity should be enforced selectively and in varying degrees over the corresponding pixels between the two images; some parts should be changed, others should stay similar. Thus, we propose to use a classifier to distinguish between images from domain $A$ and images from domain $B$, such that we extract the heat map before the classification layer to highlight the features that distinguish $A$ images from $B$ images, and require the cold part of the feature-map to stay similar. Figure \ref{fig:2} highlights the highlevel structure of our model.

\subsubsection{GAN choice}

To combat GAN instability as well as mode collapse, we choose to train a Wasserstein GAN\cite{56} instead of the original GAN. In the original GAN, the each generated instance independently is tested whether it is close to the target domain, and thus ends up minimizing the KL divergence between the generator distribution and the target distribution. The problem with minimizing the $KL[PG||PT ]$ is that when $P_G(x)$ is low, it does not matter that $P_T (x)$ is high, therefore, such loss can allow a mode collapse (ignoring intervals where $P_T (x)$ is high). This is despite the fact that the formulation for the GAN is a Jensen-Shannon loss, since PT is actually estimated by the discriminator and hinges on its quality. However, in the Wasserstein GAN\cite{56, 8} the discriminator loss is not a datapoint independent evaluation, but the batch as a whole is evaluated. This is because the Wasserstein GAN is based on a relaxation dual of the Optimal Transport solution between two point-cloud distributions, rather than continuous ones. This is more accurate in the GAN setting since access to the implicit underlying continuous distribution of both the target domain and the generator domain at anypoint of time is non-existent, instead we only have samples from each distribution. I.E., just we just have a point-cloud distribution estimation of the continuous underlying distribution. Since the whole batch/mini-batch of instances is taken into consideration, variation and diversity of modes do matter and the consequent loss reflects that. Therefore gradients can help the generator attend to unattended modes. As detailed in \cite{8}, achieving a discriminator that estimates the dual loss to the optimal transport problem requires only searching through the space of K-Lipschitz functions, therefore, they opt to clip the weights of their discriminator network.While this does indeed narrows the search space to a subset of K-Lipschitz, there is no guarantee that the best performing K-Lipschitz function lies within this subset. I.E., \cite{8} restricts the discriminator search space more than required.Meanwhile \cite{56} achieves the same property by appealing to another property: having a compact support, by penalizing the norm of the discriminator gradient with respect to random interpolated inputs between the generated samples and real samples.

\subsubsection{Architecture}

~

\noindent\textbf{Sophistication} For the problem of sophisticated-enough architectures for the discriminator, encoders and decoders, and at the same time to mitigate the problem of overfitting the training set, we initially chose DenseNets \cite{27} as the building block. DenseNets are known to allow very deep architectures while having very few number of parameters; this is because of their excellent ability to reuse previously-computed feature-maps, thus not requiring too many additional feature maps to be computed. However, since the feature-maps of one convolution are concatenated to a cumulative set of all previously computed feature-maps with the same image resolution thus far, and forwarded to subsequent layers, much higher memory consumption is ensued (copying for concatenation). Nonetheless, the authors of DenseNets actually touched on this issue, and explained that clever implementation (checkpoints) for the concatenation operator can help reduce the memory footprint dramatically\cite{44}. This required us to switch our code base from using TensorFlow\cite{2} to using PyTorch\cite{42}, since no mechanism similar to “checkpoints” is available in TensorFlow and is only resolved recently \cite{1}. We had chosen TensorFlow for its superior newly implemented data-feeding pipelines that prevent GPU starving and minimize the training time wasted on data delivery.

Unfortunately, since we opted to use the loss configuration detailed in \cite{56} for our Wasserstein GAN, as well as stability constraint detailed in \cite{41} in which both detail penalties on norms of the gradients, the PyTorch’s checkpoint functionality could not be used as it collapses gradients from the same feature-map in all concatenation places to just one gradient. This prevents the gradients from explicitly existing and therefore PyTorch’s autodiff module complains upon asking to return all gradients explicitly to compute their norm and impose the penalty (even if the collapsed gradients are the same, as is in our case). Hence, the problem of high memory consumption still persisted given the use of DenseNets. Therefore, we decided to leave DenseNets for another alternative that still provides skip-connections. ResNets\cite{25} were the obvious choice but their tendency to behave as an ensamble of shallow networks rather than just 1 deep network is concerning\cite{54}. The reason is that the residual connections summed to the output of the convolution after the non-linearity. A simple fix to the problem is to sum the skipped input to the output of the convolution before passing through the non-linearity. This is the gist of the DiracNet\cite{58}, the rest is just taking care of the dimensionality differences between the input and the output of the convolution. DiracNet resolves this problem by adding a identity convolution operator (dirac/kronecker delta window) to the convolution weights, such that when the convolution weight is applied, it also applies an identity operator implicitly. They elaborate onto this by adding learnable weights to the identity operator and others to the original convolution operator and typically normalize the original convolution operator before applying them. We use a simple version of DiracNet convolutions, were the original convolution weight is left as is, and only the identity window is weighted with learnable parameters.

\noindent\textbf{Number of parameters} To further reduce the number of parameters in our networks, we employ Depth-wise Separable Convolution \cite{13} operations in which the depth(channels) of a regular convolution filter does not expand to cover the depth (channels) of the input, but rather is thought of as having different slices of the convolution window and each slice’s sum reduces independently. This is followed by multiple 1x1 convolutions that essentially reduce the depth-separate convolution reductions in different manners. \cite{22} proved that a Depth-wise separable convolution operator learns the principal component of a regular convolution operator in its place. The major advantage is the much fewer number of parameters ($O\times W \times H\times C$) to $((O+W \times H)\times C)$, while still not skimping on the ability of modeling the convolution operator.

\noindent\textbf{Activation Functions} We used SELU activations \cite{34} throughout the architectures with the exception for the final outputs for the encoder and decoder, where we use hardtanh(0,1). SELU provide the ability to derive the output distribution of a layer to be standardized (by default to mean =0 and std = 1), without the need for a batchnorm. They also do not have the dead-ReLU problem, which is also mitigated by LeakyReLU \cite{45} and PReLU\cite{23}, but we preferred not to add more hyper parameters nor learnable ones to our system.

\noindent\textbf{InstanceNorm} Following \cite{53}, we decided to apply InstanceNorm after each convolution to preserve the sanctity of each transformed instance.

\noindent\textbf{Pooling operations} We used strided convolution to achieve resolution reduction instead of max or average pooling. This is following the results of \cite{50} in allowing the network to discover the best way the downsample a set of feature-maps.

\noindent\textbf{Convolution Transpose (Unpooling)} : As the penalty of generators (encoder and decoder) is stringent upon the quality of the images and how they match with either the input image or whether they pass as a target-looking image, it is imperative to dedicate special care for the convolution-transpose operators. Following \cite{55} and recognizing the importance and impact upsampling operations have, we implemented special conv-transpose operator that first applies Bilinear interpolation (rather than just spacing out the inputs with 0’s in-between, as done in regular ConvTranspose2d), followed by a Dirac Depth-wise separable Convolution. Not only this helps reducing the number of parameters, it also have a better input rather than non-zero dominated inputs for large strides.

\subsubsection{Model Setup}

The setup for our 1-GAN method is to train a cycle in one direction. I.e., let $x$ be a batch of image from domain $A$ and $Y$ be a batch of images from domain $B$. let $G$ be the encoder and $F$ be the decoder, $D$ the discriminator, and $C$ be the classifier: Then our loss functions are :

\begin{align*}
\begin{array}{c}
\mathcal{L}_{cyc}(x) = IHL(x - F(G(x))) \\
\mathcal{L}D(x, y) = \mathbb{E} [D(G(x))] - \mathbb{E}[D(y)] \\
\mathcal{L}_G(x) = -\mathbb{E}[D(G(x))] \\
\mathcal{L}_{\nabla_D} = ||\nabla_u \mathbb{E} [D(u)]||^p\\
\mathcal{L}_{\nabla_D}  = ||\nabla\mathcal{L}D||^2 \\
\mathcal{L}_{sim}(x) = Heatmap(C(x))\cdot||x - G(x)|| \\
\mathcal{L}_C (x, y) = log(C(x)) - log(1 - C(y)) \\
IHL(x, y) = 
\begin{cases}
||x - y||^2 & |x - y|> 1 \\
|x - y|     & |x - y|\le 1
\end{cases}
\end{array}
\end{align*}

$IHL$ is the inverted version of Huber’s loss returning the greater of $L_1$ and $L_2$ losses. $U$ in $\mathcal{L}_{\nabla_D}$ is a random linear interpolation between $y$ and $G(x)\cdot P = 6$ in our experiments following \cite{56}. The training is done in an alternating fashion; optimizing the discriminator while the encoder and the decoder are fixed, then fixing the discriminator and optimizing the encoder and the decoder. The Classifier is trained alongside both alternates. Listing 1 describes the training procedure for the 1-GAN model.

\begin{algorithm}[hbt!]
\caption{ 1-GAN Training}\label{alg:1}
\begin{algorithmic}
\State \textbf{procedure} 1-GAN
\State \quad $x, y \gets $ minibatches from domains $A$ and $B$ respectively
\State \textbf{while} not converged \textbf{do}
\State \quad \textbf{with} $G$ and $F$ fixed:
\State \quad \quad Minimize $\mathcal{L}_D(x, y) + 2\mathcal{L}_{\nabla_D}$
\State \quad \textbf{if} $\mathcal{L}_D(x, y)<0$ \textbf{then} \Comment{$D$ can differentiate between generated and target}
    \State \quad\quad \textbf{with} $D$ fixed:
    \State \quad\quad\quad Minimize $\mathcal{L}_{cyc} + \mathcal{L}_G + \frac{1}{2} \mathcal{L}_{\nabla_G} + \mathcal{L}_{sim}$  
\State \quad Minimize $\mathcal{L}_C$
\end{algorithmic}
\end{algorithm}

\subsection{The GAN-free Model}

Variational autoencoders\cite{33} are generative models that map the instances from the input domain to an explicit probability distribution (a prior), such that samples from the prior distribution should be able to construct the input instance again, as well as, generate domain-looking unseen instances. Typically the choice of the prior distribution is a multidimensional standard Gaussian, because of its ability to approximate any sophisticated distribution when pushed through a sophisticated-enough deterministic function. Our intention here is to employ two variational autoencoders to represent the source domain and the target domain respectively using the same prior distribution, and then enforce a cycle consistency constraint to perform the mapping.

Therefore the problem becomes: given an image from domain $A$, pick out a representation from the prior such that it can construct image $B^{'}$, where $B^{'}$ picks out a representation from the prior that can construct image $A$ back. The main reasoning here is that an image and its representation are tightly coupled, such that if $A$ wants to be reconstructed back at the end, $A$’s representation has to keep enough information of the input $A$, and as a byproduct, this information constructs $B^{'}$ which is similar to $A$ but from the other domain.

\subsubsection{Architecture}

We used the same set of architectual building blocks detailed in the 1-GAN formulation, with the notable exception that a encoder is a mapping from an image to a latent vector and a decoder maps from a latent vector to an image.

\subsubsection{Autoencoder choice}
~

\noindent\textbf{Variational Autoencoders} The convolutional version of the original variational autoencoder (VAE) maps images from the image domain each to a parameters of a multidimensional Gaussian (reparameterization trick) such that if a vector is sampled from this Gaussian and passed through the decoder, the original image is reconstructed. Therefore it maps every image to a local Gaussian neighborhood that can reconstruct that image. This alone is not enough to constrain the encoder not to overfit and just assign Guassian means that are far apart with variances that are small. So the original variational autoencoder adds another distributional (prior) constraint: the local neighborhoods should all be converging to the standard normal Gaussian $\mathcal{N} (0, I)$, forcing these local neighborhoods to be close to each other such that you can interpolate from one neighborhood to another. Therefore sampling from the mulivariate standard gaussian and passing through the decoder can generate images that look like the original dataset. Traditionally, the prior constraint is a $\mathbf{KL}$ divergence term pushing each neighborhood to be more like $\mathcal{N} (0, I)$, and the reconstruction constraint is either an MSE loss or a binary-cross-entropy/negative-log-likelihood. These two losses are added together and minimized simultaneously, unlike GANs which require alternating minimization because of the minimax game setup.

\noindent\textbf{$\mathbf{\beta}$-VAE} However, the issue of neighborhood closeness is of much concern; it has been inadvertently studied in the context of disentangled representations\cite{39, 9}. \cite{4} studied the problem from an information bottleneck point of view and verified the findings of \cite{39}; they formulated the problem as a distortion-rate trade off, where the reconstruction loss is the distortion and the divergence from the prior is the rate. They revealed that since the total entropy of the dataset is fixed, the distortion and rate are at a trade off ( one cannot minimize both arbitrarily). Essentially, if the image’s vector representation carries all the information about the image (no loss) then it must diverge from the prior as there is little chance the image domain is completely standard Guassian-looking. And if you want the representation vector to adhere to the prior, you must be losing some information about the particular image. The sweet spot where the lost information is not much such that you can still construct a similar looking image, while your representation adheres to the prior fairly enough. In other words, the neighborhoods are each not too divergent from $\mathcal{N} (0, I)$ while the reconstructed image does not suffer alot of dissimilarity loss to the original. This is enforced by having a modulating hyperparamterer $\beta$ to tune the emphasis between the reconstruction loss and the prior adherence loss.

\noindent\textbf{VampPriors\cite{52}} Since the ultimate theoretical prior is one that highlights all the “good” representations given all the images in the dataset (ie, aggregate prior), and to alleviate the problem of adherence to a static prior of $\mathcal{N} (0, I)$, VampPrior VAE \cite{52} proposes to learn the prior itself by assuming that the total prior is just a mixture of posteriors of K synthetic images. The synthetic images themselves are backpropaged and adjusted, and thus the prior distribution is adjusted. The synthetic images form probabilistic basis for all the images in the dataset. When sampling to generate unseen images, one would sample not from $\mathcal{N} (0, I)$ but rather from this prior mixture of the synthetic images, therefore eliminating parts of the latent vector space that is not covered by a basis neighborhood.

\noindent\textbf{Sinkhorn Autoencoders} The search for a powerful and flexible way to make latent representations adhere to any sophisticated and explicit prior is one of the active research areas. Sinkhorn Autoencoders \cite{43} employ recent advancements in the Optimal Transport theory between point-cloud distributions to help make more powerful autoencoders. \cite{15} shows a deterministic algorithm used to compute an entropy-regularized version of the relaxed primal optimal transport problem, thus giving approximate but accurate optimal transport loss. Since the algorithm is nothing but a linear program to a constraint optimization problem that can be expressed in fairly simple matrix-algebra operations, it can be implemented using operations that are supported by automatic differentiation packages, and thus one can obtain gradients to minimize or maximize the computed loss. This opened the opportunity for many probabilistic models to try to utilize such tool in order to enforce distribution matching without assumptions on the type or shape of the distributions in question. An immediate application is the Sinkhorn autoencoders, where samples from encoded representations are matched against samples from the desired prior distribution and the sinkhorn loss is minimized using backpropagation. One draw back is that the first step of computing the sinkhorn loss requires pairwise comparisons between samples from one distribution against the other. The usual comparison (cost) functions are L1 or L2 losses, thus they can suffer from the curse of dimensionality if the compared vectors are of high dimensions.\footnote{We actually contacted the first author of the Sinkhorn Autoencoders regarding this and other instabilities stemming from the fact that the loss is entropy regularized, and the cost matrix can quickly die when exp(- CostMatrix) is performed.It appears that for the current formulations there is no escaping that, thus one should use a manageable latent vector dimensionality} Moreover, L1 and L2 are not valid comparisons for images since a simple shift of pixels can result in very high loss, while perceptually they still look identical. Recognizing this, \cite{19, 48} independently suggested the use of a learnable function $\phi$ to map the highly dimensional to-be-compared vectors to a low-dimensional space where L1 or L2 differences are applicable. However, in the translation problem (latent vector to image, or, image to image), training the untrained decoder/generator to produce target-domain-looking images one requires finding a map $\phi$ that first highlights the differences between the current generated images and the target images, then, optimize the decoder/generator to minimize the loss. Which formulates a minimax problem.

In our experiments we used Beta varational auto encoders and later Sinkhorn autoencoders skipping VampPriors because of the latter’s flexibility and VampPriors’s dependence on the choice of number of synthetic images and the extensive amount of computation they require. We follow \cite{48} in using the normalized sinkhorn loss as well as using the cosine dissimilarity to as the cost between sinkhorns.

\subsubsection{Model Setup}

Here we tried many different model configurations, and we will detail 4 prominent ones, and discuss their results in the experiments and discussion sections:

\noindent\textbf{The Sequential $\mathbf{\beta}$-VAE} In this configuration we had two $\beta$-VAE one modeling each image domain. I.e. $\beta$-VAEA would apply reconstruction loss and prior loss on images from domain $A$. Similarly $\beta$-VAEB does the same for images from domain $B$. Moreover, $\beta$-VAEA would accept images from the $B$ domain but only enforce a prior adherence loss on them. $\beta-\mathrm{VAE}_B$ also accepts $A$ images and enforces prior adherence loss on them. For the last two cases, the unenforced reconstruction loss is replaced by a cycle reconstruction loss. In other words, while $B$ images are required to adhere to $\beta-\mathrm{VAE}_A$ ’s prior (which is identical to $\beta-\mathrm{VAE}_B$’s prior), the output of $\beta-\mathrm{VAE}_A$’s decoder is passed into $\beta-\mathrm{VAE}_B$ where the output of the latter has to be the exact $B$ input image to $\beta-\mathrm{VAE}_A$.

Therefore the losses here are:

Self-domain losses:
\begin{align*}
\begin{array}{l}
\mathcal{L}_{V~AE_A} = IHL(x - D_A(E_A(x))) + \beta KL[E_A(x)||\mathcal{N} (0, I)] \\
\mathcal{L}_{V~AE_B} = IHL(y - D_B(E_B(y))) + \beta KL[E_B(y)||\mathcal{N} (0, I)]
\end{array}
\end{align*}

\noindent Cycle-losses:
\begin{align*}
\begin{array}{l}
\mathcal{L}_{CY C_A} = IHL(x - D_A(E_A(D_B(E_B(x))))) \\
+ \beta KL[E_B(x)||\mathcal{N} (0, I)] + \beta KL[E_A(D_B(E_B(x))))||\mathcal{N} (0, I)]
\end{array} 
\end{align*}
\begin{align*}
\begin{array}{l}
\mathcal{L}_{CY C_B} = IHL(y - D_B(E_B(D_A(E_A(y))))) \\
+ \beta KL[E_A(y)||\mathcal{N} (0, I)] + \beta KL[E_B(D_A(E_A(y))))||\mathcal{N} (0, I)]
\end{array}
\end{align*}

\begin{figure}
    \centering
    \includegraphics[width=0.48\textwidth]{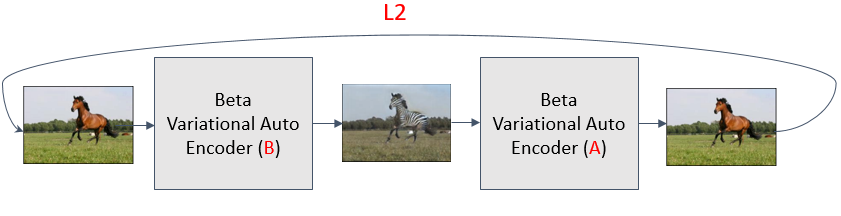}
    \caption{The Sequential $\beta$-VAE model}
    \label{fig:3}
\end{figure}

The idea here is that both encoders should recognize images from the opposite domain and produce representations that adhere to the prior such that their own domain decoder produces an image close in characteristics enough to carry out the reconstruction at the end of the cycle (once it passes through the other VAE). Figure \ref{fig:3} shows the structure of this model.

A more strict variant of this configuration is also implemented by freezing (fixing) the decoders during the cycle loss optimization. This restricts the decoders from changing themselves from how they decode the prior just to accommodate the translated image. This way the problem is turned into optimizing the encoders to pick the right representations that lead the decoders to the right translation and reconstruction.

Interleaving $\beta$-VAEs In this configuration one domain’s $\beta$-VAE is flipped and sandwiched between the other domain’s $\beta$-VAE’s encoder and decoder. For example, a cycle is carried out as follows $E_A \to D_B \to E_B \to D_A$. Here the autoencoder for domain $B$ is flipped (decoder first) and sandwiched between domain $A$’s autoencoder. Unlike the previous model, here it is a decoding problem than an encoding one. Each encoder accepts only images from its own domain, it is up to the decoders to learn to decode the other domain’s image’s representation properly to minimize the cycle reconstruction loss (reconstructing the same input image at the end). Therefore in addition to the regular self-domain $\beta$-VAE losses ($\mathcal{L}_{V~AE_A}, \mathcal{L}_{V~AE_B}$ ) in the previous model the cycle losses here are:

\begin{align*}
\begin{array}{l}
\mathcal{L}_{CY C_A} = IHL(x - D_A(E_B(D_B(E_A(x))))) \\
+ \beta KL[E_A(x)||\mathcal{N} (0, I)] + \beta KL[E_B(D_B(E_A(x))))||\mathcal{N} (0, I)]
\end{array}
\end{align*}
\begin{align*}
\begin{array}{l}
\mathcal{L}_{CY C_B} = IHL(y - D_B(E_A(D_A(E_B(y))))) \\
+ \beta KL[E_B(y)||\mathcal{N} (0, I)] + \beta KL[E_A(D_A(E_B(y))))||\mathcal{N} (0, I)] 
\end{array}
\end{align*}

\begin{figure}
    \centering
    \includegraphics[width=0.48\textwidth]{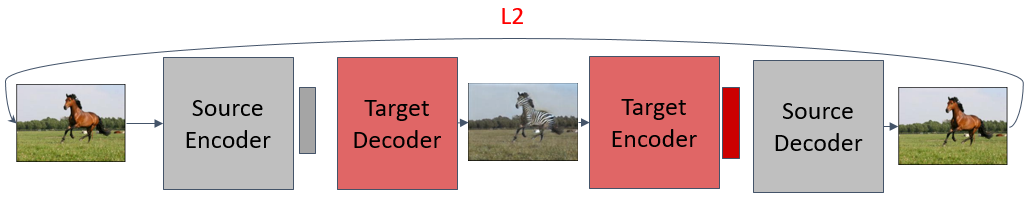}
    \caption{The Interleaving $\beta$-VAE model}
    \label{fig:4}
\end{figure}

This model configuration is visually depicted in Figure \ref{fig:4}. Also another stricter version of this was

tried where the encoders were frozen in the to disallow any changes to the way the images are encoded by their own domain’s VAE just to accommodate the cycle loss.

\textbf{Aligned encoding} This configuration has a strict assumption that none of the encoders or decoders should change to accommodate the cycle loss. It frames the problem into pure representation aligning problem. The $\beta$-VAE of both domains are trained on their respective domains without cross-domain constraints. The cycle loss is enforced by having a Alignment (translation) network that translates a latent vector representation from one domain to another such that the cycle loss is minimized. An alignment network is a simple two layer MultiLayerPerceptron (MLP), denoted $ALGN_{B2A}$ and $ALGN_{A2B}$ below for each alignment direction. So in addition to self-domain VAE losses ($\mathcal{L}_{V AE_A}, \mathcal{L}_{V AE_B}$), cycle losses :

\begin{equation*}
\begin{split}
\mathcal{L}_{CY C_A} = IHL(x - D_A(ALGN_{B2A}(E_B(D_B(ALGN_{A2B}\\ 
(E_A(X))))))) \\
\mathcal{L}{CY C_B} = IHL(y - D_B(ALGN_{A2B}(E_A(D_A(ALGN_{B2A}\\
(E_B(Y )))))))
\end{split}
\end{equation*}

Moreover, prior adherence losses are imposed on the output of the alignment networks, as well as representation-mirror constraints; $ALGN_{A2B}$’s input and output should also mirror $ALGN_{B2A}$’s outputs and inputs, and vice versa. In other words,

\begin{equation*}
\begin{split}
\mathcal{L}_{algnprior_A} = KL[ALGN_{A2B}(Z_A)||N (0, I)] \\
\mathcal{L}_{algnprior_B} = KL[ALGN_{B2A}(Z_B)||N (0, I)] \\
\mathcal{L}_{mirror_A} = L2(Z_A - ALGN_{B2A}(ALGN_{A2B}(Z_A))) \\
\mathcal{L}_{mirror_B} = L2(Z_B - ALGN_{A2B}(ALGN_{B2A}(Z_B)))
\end{split}
\end{equation*}

Where $Z_A$ and $Z_B$ are any representation inputs to the align networks during the training.

\begin{figure}
    \centering
    \includegraphics[width=0.48\textwidth]{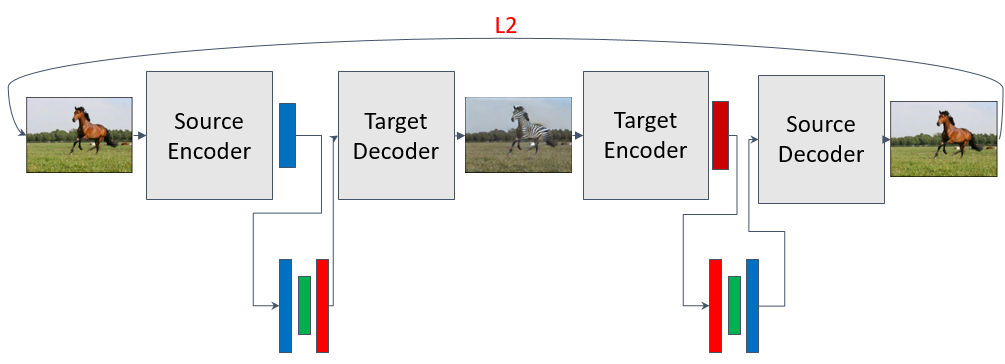}
    \caption{The Aligned $\beta$-VAE encoding model}
    \label{fig:5}
\end{figure}

It is worth noting that during enforcing the cycle consistency training, $\beta-\mathrm{VAE}_A$ and $\beta-\mathrm{VAE}_B$ are frozen such that only the alignment networks are being optimized. The model is detailed in Figure \ref{fig:5}.

\begin{figure}[ht]
    \centering
    \includegraphics[width=0.48\textwidth]{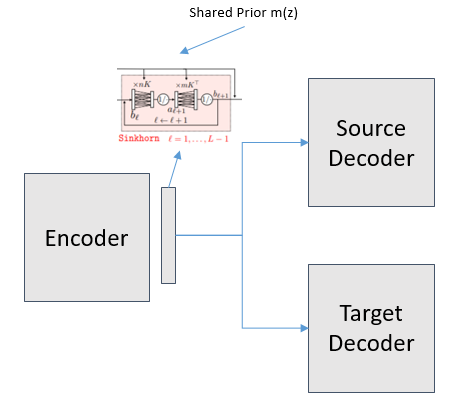}
    \caption{The Sinkhorn shared encoder model}
    \label{fig:6}
\end{figure}

\textbf{Sinkhorn shared encoder} To to enforce a shared latent space and shared feature extraction, we resolved to using one encoder for both domains. However, such decision requires have a powerful prior adherence constraint without imposing neighborhood shape assumptions. Therefore it is only appropriate to use something as powerful and flexible as a sinkhorn loss to align the point-cloud distributions between the generated representations and the desired prior.

\begin{equation*}
\begin{split}
\mathcal{L}_{SAE_A} = IHL(x - D_A(E(x))) + \beta \mathcal{L}_{sink}(E(x), z) \\
\mathcal{L}_{SAE_B} = IHL(y - D_B(E(y))) + \beta \mathcal{L}_{sink}(E(y), z)
\end{split}
\end{equation*}

\begin{equation*}
\begin{split}
\mathcal{L}_{CY C_A} = IHL(x - D_A(E(D_B(E(x))))) + \beta \mathcal{L}_{sink}(E(x), z) \\
+ \beta \mathcal{L}_{sink}(E(D_B(E(X)))), z)\\
\mathcal{L}_{CY C_B} = IHL(y - D_B(E(D_A(E(y))))) + \beta\mathcal{L}_{sink}(E(y), z) \\
+ \beta\mathcal{L}_{sink}(E(D_B(E(Y )))), z)
\end{split}
\end{equation*}
Where $\mathcal{L}_{sink}$ is the normalized sinkhorn loss given by $\mathcal{L}_{sink}(R, Q)  = 2sinkhornLoss(R, Q) - sinkhornLoss(R, R) - sinkhornLoss(Q, Q)$ This is akin to mutual loss of information given two point-cloud distributions. Figure \ref{fig:6} details this model.

\section{Experiments and Results}

\subsection{Datasets}

In order to carry out initial hypothesis testing, we used the MNIST dataset as one domain, and synthetically morphed it to come up with target domains. Among the operators applied were: color inversion, rotation, horizontal and vertical flipping and stretching. The intention was that once we confirm that our models work on such a toy dataset, we start testing on benchmarking datasets that were tried in CycleGAN\cite{60} and pix2pix\cite{29} like Cityscapes\cite{14} and imagenet’s\cite{16} horse and zebra classes, in order to numerically and empirically verify the capabilities of our models, respectively.

\subsection{The 1-GAN model}

We started by the constructing the architecture as in the detailed in section but with a standard ConvTranspose2d operation. However, we also added U-Net\cite{46}-like skip-connections from the downward path towards the corresponding resolution in the upward path. We quickly came to realize how useless a static classifier for the heatmap would be, since for our toy dataset, a heatmap would always just be the majority of background pixels as an indicator of the class, without even considering the shape of the digit.

\begin{figure}[ht]
    \centering
    \includegraphics[width=0.4\textwidth]{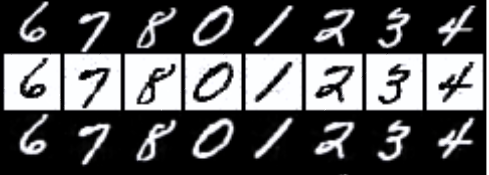}
    \caption{The initial trial for color inversion, with U-NET skip-connections(Top row: source, middle row: translation, last row: reconstruction of input from translation)}
    \label{fig:7}
\end{figure}

\begin{figure}
    \centering
    \includegraphics[width=0.15\textwidth]{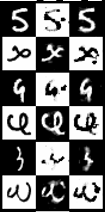}
    \caption{The initial trial for color inversion+ rotation 90-degrees + horizontal flip , without U-NET skip-connections(left col: source, col row: translation, right col: reconstruction of input from translation)}
    \label{fig:8}
\end{figure}

For our first trial, we tried on MNIST and it’s color inverse Fig as our two domains. The results were perfect (Fig \ref{fig:7}). In fact, too perfect to the point of raising suspicions. Therefore we tried a harder objective: we added a horizontal flip and a 90-degree rotation to the target domain in addition to the color inversion. Our suspicions were in place, as the model would either not honor the rotation nor the horizontal flip, and all the translated images were at best a color inversion (Fig \ref{fig:8}).

We removed the U-Net skip connections to find that the network struggles to output anything of resemblance to the target domain save the color of the background. This triggered over a month-long of constant architecture, configuration and hyperparameter testing. We tried adjusting the weight initialization schemes since in Dirac Convolutions there are essentially two outputs summed before the non-linearity, so we adjusted the gain in for He initialization\cite{23} and tried glorot\cite{20} initialization since it accounts for the fan-in as well. We also tried the truncated Normal initialization as SELU\cite{34} suggests. None of those tweaks resulted in producing any successful training. We removed the Dirac configuration, and later also the depthwise separable convolution. We then stripped it down to regular convolutions and deconvolutions. Replaced SELUs with LeakyReLUs and then ReLUs. We enabled biases, disabled biases. We added regular residual connections like ResNets thinking it would help reduce the effective depth of the network.Given none of the aforementioned and many many other architectural tweaks produced any working-resembling models, we turned our attention to the choice of GAN, and decided to try the original Wasserstein GAN\cite{8} (with weight clipping), and that still did not give any nudging results. The strange part is that most of the aforementioned tweaks were also tested by us and proved to be working for tasks like multi-class classification, auto-encoding, variational auto-encoding and even regular GAN training (noise to image generation), so we assumed that a conditional GAN training (image to image) has its own sensitivity since the generator is implicitly doing two parts (image to implicit representation , implicit representation to image). At this point we assumed that the hyperparameter tuning for the optimizer, especially the learning rate, has proved to be highly sensitive for the conditional GAN setting and that we just have not come to any valid selections of learning rates, weight decay, and momentum, given any of the model and GAN configurations we tried, and the many optimizers we configured. We set out to implement Adasecant\cite{62}. Which boasts a learning rate-free optimization by estimating different learning rates for each direction of the gradient after estimating the local hessian using finite difference and directional newton methods. After we implemented and verified the adasecant optimizer for PyTorch on various simple models (classification ,autoencoding, noise-to-image GAN). We tried it on our image-translation configuration but to no avail; the cycle loss would not go down after certain still-high threshold. Thinking that it might be a saddle point problem\cite{17}, we also incorporated gradient perturbation detailed in \cite{30, 31}, but that did not help either.

It was not until we re-implemented the models multiple times with careful gradient tracing that we got our models to train. It was a combination of things, but mainly having to give special care to PyTorch gradient manipulation and properly detaching some gradients from the auto-differentiation graph and keeping others registered while passing through frozen modules.

\begin{figure}
    \centering
    \begin{subfigure}{0.24\textwidth}
    \centering
        \includegraphics[width=0.5\textwidth,height=4.5cm]{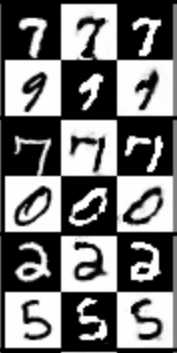}
        \caption{color-inversion\\~}
        \label{Fig:9a}
    \end{subfigure}
    \begin{subfigure}{0.24\textwidth}
    \centering
        \includegraphics[width=0.5\textwidth,height=4.5cm]{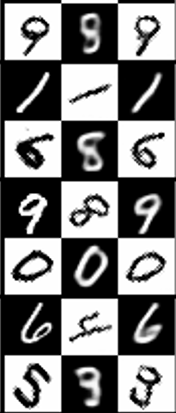}
        \caption{ color-inversion + rotation 15-degrees}
        \label{Fig:9b}
    \end{subfigure}
    \caption{ Results for the working model on MNIST dataset pairs, result in (a) are after only 2 epochs}
    \label{fig:9}
\end{figure}

We reverted back to our initial architectures with a reworked Conv Transpose/Unpooling operator as detailed in the architecture section, and verified that the model was heading in the right direction on the color-inversion task by itself without U-net skip-connections (Fig \ref{Fig:9a}). We then applied it to color-inversion + rotation and it produced satisfactory results (Fig \ref{Fig:9b}). We mostly used an Adadelta optimizer assisted with a Nesterov momentum for fast initial convergence.

It is clear that the translation is indeed is from the target domain (color-inverted rotated images), however, the mapping is arbitrary. This is acknowledged in the literature as the problem if unsupervised unpaired image-to-image translation is under-constrained, and given the sophisticated capabilities of deep neural networks, highly-convoluted mappings that still satisfy the cycle consistency can exist\cite{6, 5}. The problem is generally known as manifold alignment problem.

\begin{figure}
    \centering
    \includegraphics[width=0.48\textwidth]{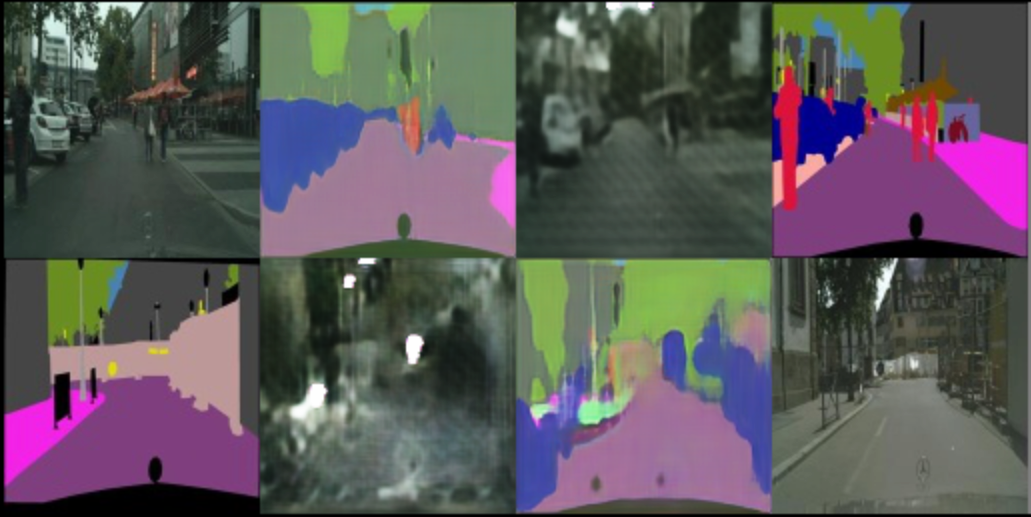}
    \caption{Training the 1-GAN model on the Cityscapes dataset with a very large learning rate. Top row: Source image -> translated image -> reconstructed source image and Translation ground truth for comparison. Bottom row: a target image -> translated to a source image -> reconstructed target image and source ground truth for comparisons. We only explicitly train for the top-row cycle direction}
    \label{fig:10}
\end{figure}

We then set the infrastructure to train our model on Cityscapes, but due to time constraint we had to train with a very large learning rate such that we see quickly if the model is learning anything even though it would not be able to converge to a good minima due to oscillations by a large learning rate. The initial results (leveled off after 10 epochs) are depicted in Fig \ref{fig:10}. It is obvious that the model is heading towards mapping the translation to the proper domain. We note here that the trained cycle real->colored\_masks (encoder -> decoder) is showing good results that appear heading into the right direction, but reverse cycle (decoder-> encoder) is producing bad results. But we argue this is because that the decoder is only conditioned to produce results based on the output of the encoder, and not on the images from the target domain (colored masks). And we argue that when training for the forward cycle is producing accurate enough target images, the reverse cycle will be producing good results as well. Just as the reverse cycle in the previous MNIST translation was producing good results.

The final architecture of a generator/encoder we settled on was:

downwards: $3conv64,~pool,~3conv128,~pool,~3conv256$.

upwards: $1deconv256,~1conv256,~1deconv128,~1conv.$

Where RconvL represents R layers of Dirac Depth-wise Convolutions with L feature-maps followed by InstanceNorm and SELU. The R are all of kernel size of $3\times 3$ layers are dialted differently $3-2-1$ \cite{10}.

Pooling operators are 1convL but with strides of 2, where $L$ is the same as the upcoming layer’s $L$.

$RdeconvL$ represents a Dirac Depth-wise Convolutions with $L$ feature-maps after a Bilinear upsampling\cite{55} of factor-size of 2. All kernel sizes for $deconv$ were 4 with stride 2.

A discriminator/classifier has the exact first part of the downard architecture with a reduction to dimenions of $Zx1x1$ with appropriate kernel size to bring it down a resolution of $1x1$ (depending on the image resolution), $Z$ is 10 in all our experiments for the discriminators, $Z$ is 100 for autoencoders: $3conv64$, $pool$, $3conv128$, $pool$, $3conv256$, $1convZ$.

For the Wasserstein discriminator, the output is the mean of $Zx1x1$ at the end. 

With these configurations, a generator/encoder’s size was only 480K parameters and a discriminator’s size 322K parameters, compared to CycleGANs 11.37M and 2.76M parameters for a generator and a discriminator respectively.

\subsection{GAN free models}

\subsubsection{The sequential $\beta$-VAE model}

\begin{figure}
    \centering
    \includegraphics[width=0.4\textwidth]{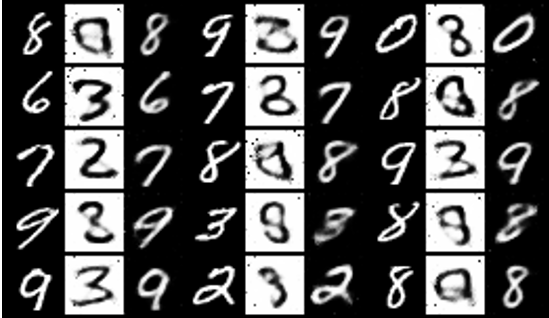}
    \caption{Results for the Sequential $\beta$-VAE models: Good reconstruction but arbitrary mappings}
    \label{fig:11}
\end{figure}

The sequential $\beta$-VAE model was able to achieve good cycle reconstruction but arbitrary mappings, Fig \ref{fig:11}. This is also an instance of manifold alignment. Some of the digits are valid while others are obscure. Since the prior adherence is enforced, these problems are due the existence of regions of the prior that the was never trained on and do not strongly belong to the neighborhoods of any valid instances. Similar results were achieved when applying the strict variant of the sequential $\beta$-VAE.

\begin{figure}
    \centering
    \includegraphics[width=0.4\textwidth]{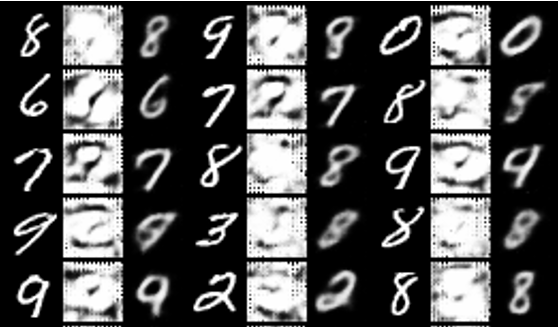}
    \caption{Results for the Interleaving $\beta$-VAE models: Again passable reconstruction but arbitrary mappings very much suffering from prior holes doubly}
    \label{fig:12}
\end{figure}

\subsubsection{The interleaving $\beta$-VAE model}

The interleaving $\beta$-VAE model and its variants also resulted in good cycle reconstruction for the most part while maintaining arbitrary mapping and mostly due to holes in the prior (Fig \ref{fig:12}). We tried enforcing one more constraint by having the translated image be encoded by its domain encoder and decoded by a separated trained domain encoder and requiring the input translation image to be equal to the resulting decoded image (by L2). This is to enforce a one-to-one mapping to encoded representations of the encoder. But this did not produce good results either since the separate decoder can still suffer from the prior holes as well.

\begin{figure}
    \centering
    \includegraphics[width=0.4\textwidth]{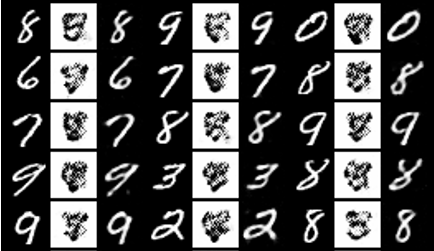}
    \caption{Results for the Alginment Networks models: strict adherence to the prior holes}
    \label{fig:13}
\end{figure}

\subsubsection{The Alignment Networks model}

The Alignment Networks model strictly produced mappings that are a mishmash of various neighborhoods (holes in the prior but mostly adherent) (Fig \ref{fig:13}).

\subsubsection{The Sinkhorn Shared Encoder}

\begin{figure}[ht]
    \centering
    \includegraphics[width=0.15\textwidth]{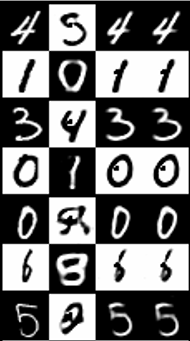}
    \caption{Results for the Sinkhorn Shared encoder Networks models: strict adherence to the prior, barely suffering from holes, but full manifold alignment manifestation. Image -> Translation -> reconstruction and its direct Autoencoding for comparison. Middle black dot was an accidentally added artifact while saving the image samples}
    \label{fig:14}
\end{figure}

Finally, the sinkhorn autoencoding scheme produced satisfactory mappings and cycle reconstructions but again the manifold alignment problem still manifested (Fig \ref{fig:14}). It is also noted that the sinkhorn decoders still suffered from prior holes but in much less pronounced fashion than $\beta$-VAE models.

\begin{figure*}
    \centering
    \includegraphics[width=\textwidth]{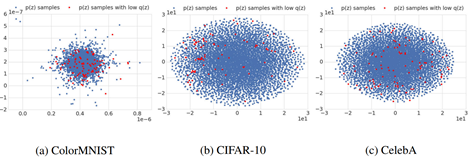}
    \caption{Image from \cite{47}: Examples of prior holes on variationally autoencoded datasets, red dots represent pockets that the decoder does not know how to decode properly and do not result in good output}
    \label{fig:15}
\end{figure*}

\section{Discussion and Future work}

\subsection{Benchmarks}

Unfortunately we could not have reported any benchmarking comparisons due to time constraints, and we only evaluated our models empirically. We will be working a full evaluation in the near future. In addition, further improvements to the ConvTranspose operator are needed to reduce the chalkboard artifacts apparent at the beginning of the training and hinting that more training might be straining the deconvolution operators\cite{51}. This might help us achieve even more smoother training.

\subsection{Manifold Alignment}

Due to the nature of the unsupervised unpaired image-to-image translation being ill-posed; not enough constraints dictating how the translation should go, arbitrary but valid mapping keep manifesting in our models. This is because the cycle consistency is only a weak constraint. This is verified by the ability to satisfy a the cycle reconstruction in all our tests while passing through arbitrary mapping from the target domain. This is, unsurprisingly, due to the highly-sophisticated mappings that can be achieved by deep neural networks. CycleGAN achieves translations similar to the input image only because of locality of the convolution operator and that the target domain patterns are also functions of local parts of the image. Thus in the eyes of the optimizer, it is easier to change local parts of the image to achieve good discriminator loss than to completely conjure up a new image from the target domain. This is also verified by our findings on the MNIST dataset pairs, when rotated (ie no straight digits in the target domain) the only way to satisfy the domain translation is to generate a rotated color-inverted digit, and in doing so, the locality of convolution bias is no longer acting and therefore, any such rotated digit satisfy the mapping. Whereas when only color inversion is required (local change) the mapping was exactly the same digit but with inverted colors. This is also verified by the our generated cityscape translations, they are not arbitrary, and it is empirically evident that local features are responsible for features in the generated masks and for the reconstruction features.

The manifold alignment problem is more evident in the GAN-free models as all representations are reduced down to a latent vector, thus all and any convolution locality is lost.

\subsection{Prior holes}

Another problem prevailing in auto-encoding models is that there is no handle on the picked up translation representations other than the cycle loss, which is, again, a weak constraint in the realm of deep neural networks. The presence of holes in the decoder’s understanding of the prior can produce not only arbitrary mappings but also mappings that are not even from the target domain, rather just noisy images. This has been heavily studied by \cite{47} and previously addressed by \cite{40}, both suggest to employ a discriminator fill the holes with meaningful decodings. Other models BicycleGAN\cite{61}, UNIT\cite{36} and MUNIT\cite{28} were forced to employ a discriminator such that the decoder is conditioned to only produce target-looking images anyhow.

\subsection{Conclusions and future work}

\begin{itemize}
\item 
We have achieved a working an image translation system with fraction of the number of parameters than used by CycleGAN. Although not completely verified in a benchmarking manner, we are certain that we can achieve similar results or even surpass CycleGAN on benchmarking datasets that we will verify in the very near future.

\item 
We have verified that the unsupervised unpaired image-to-image translation problem is under-constrained and ill-posed. Our initial proposal of utilizing a classifier turned out to be naive since it can stick to superficial features, but further improvements to the mechanism can be viable to add more constraints, by generating a pixel-wise binary classification mask and apply it onto the translated image rather than extracting a heatmap from an image-wide class.
\item 
A complete manifestation of the manifold alignment problem and the prior holes problem as well as the lossy compression nature of reducing images to representations is preventing probabilistic autoencoding schemes from being a viable unsupervised image-to-image translation infrastructure. Future works should try to eliminate prior holes possibly by enforcing a VampPrior scheme where samples from trusted neighborhoods are allowed while using a powerful prior adherence loss such as sinkhorns. Attention mechanisms or the previously mentioned pixel-wise classification scheme can help bring constrain the problem locally to further itself away from a complete arbitrary manifold alignment. Lastly, instead of autoencoding images to representations, we could try encoding translation operators themselves, and apply the operator on the images. This is akin to running a skip connection from the input to the output of the autoencoder. However, as this adds a shortcut to the reconstruction loss, a self-reconstruction of the image is no longer viable, but only a cycle reconstruction. A better autoencoding scheme might also be by involving an attention mechanism to only distill the object of interest or the pattern of interest before passing through the probabilistic autoencoder, that that reconstruction loss is also contingent on the attention. Feature disentanglement might be able to help the autoencoder not be gullible in trying to encode the entire passed object.
\end{itemize}

\bibliographystyle{IEEEtran}
\bibliography{references}

\end{document}